\documentclass{bmvc2k}
\usepackage{ctable} 
\usepackage{array, makecell} %
\usepackage{tabularx,booktabs}
\usepackage{multirow}
\usepackage{amssymb} 
\usepackage{pifont}
\usepackage{calrsfs}
\usepackage{svg}
\DeclareMathAlphabet{\pazocal}{OMS}{zplm}{m}{n}
\usepackage[final]{pdfpages}
\usepackage{pifont}

\newcommand{\xmark}{\ding{55}}
\newcommand{\expnumber}[2]{{#1}\mathrm{e}{#2}}

\title{NOD: Taking a Closer Look at Detection under Extreme Low-Light Conditions with Night Object Detection Dataset}

\addauthor{Igor Morawski}{ }{1}
\addauthor{Yu-An Chen}{ }{2}
\addauthor{Yu-Sheng Lin}{ }{3}
\addauthor{Winston H. Hsu}{ }{4}

\addinstitution{
 National Taiwan University
}

\runninghead{Morawski, Chen, Lin, Hsu}{NOD (Night Object Detection) Dataset}

\begin{document}

\maketitle

\begin{abstract}Recent work indicates that, besides being a challenge in producing perceptually pleasing images, low light proves more difficult for machine cognition than previously thought. In our work, we take a closer look at object detection in low light. First, to support the development and evaluation of new methods in this domain, we present a high-quality large-scale Night Object Detection (NOD) dataset showing dynamic scenes captured on the streets at night. Next, we directly link the lighting conditions to perceptual difficulty and identify what makes low light problematic for machine cognition. Accordingly, we provide instance-level annotation for a subset of the dataset for an in-depth evaluation of future methods. We also present an analysis of the baseline model performance to highlight opportunities for future research and show that low light is a non-trivial problem that requires special attention from the researchers. Further, to address the issues caused by low light, we propose to incorporate an image enhancement module into the object detection framework and two novel data augmentation techniques. Our image enhancement module is trained under the guidance of the object detector to learn image representation optimal for machine cognition rather than for the human visual system. Finally, experimental results confirm that the proposed method shows consistent improvement of the performance on low-light datasets.  
\end{abstract}

\begin{figure}
\begin{center}
\vspace{0.5em} 
\begin{tabular}{c l}
\multirow{2}{*}[5.2em]{ 
\includegraphics[width=5.55cm]{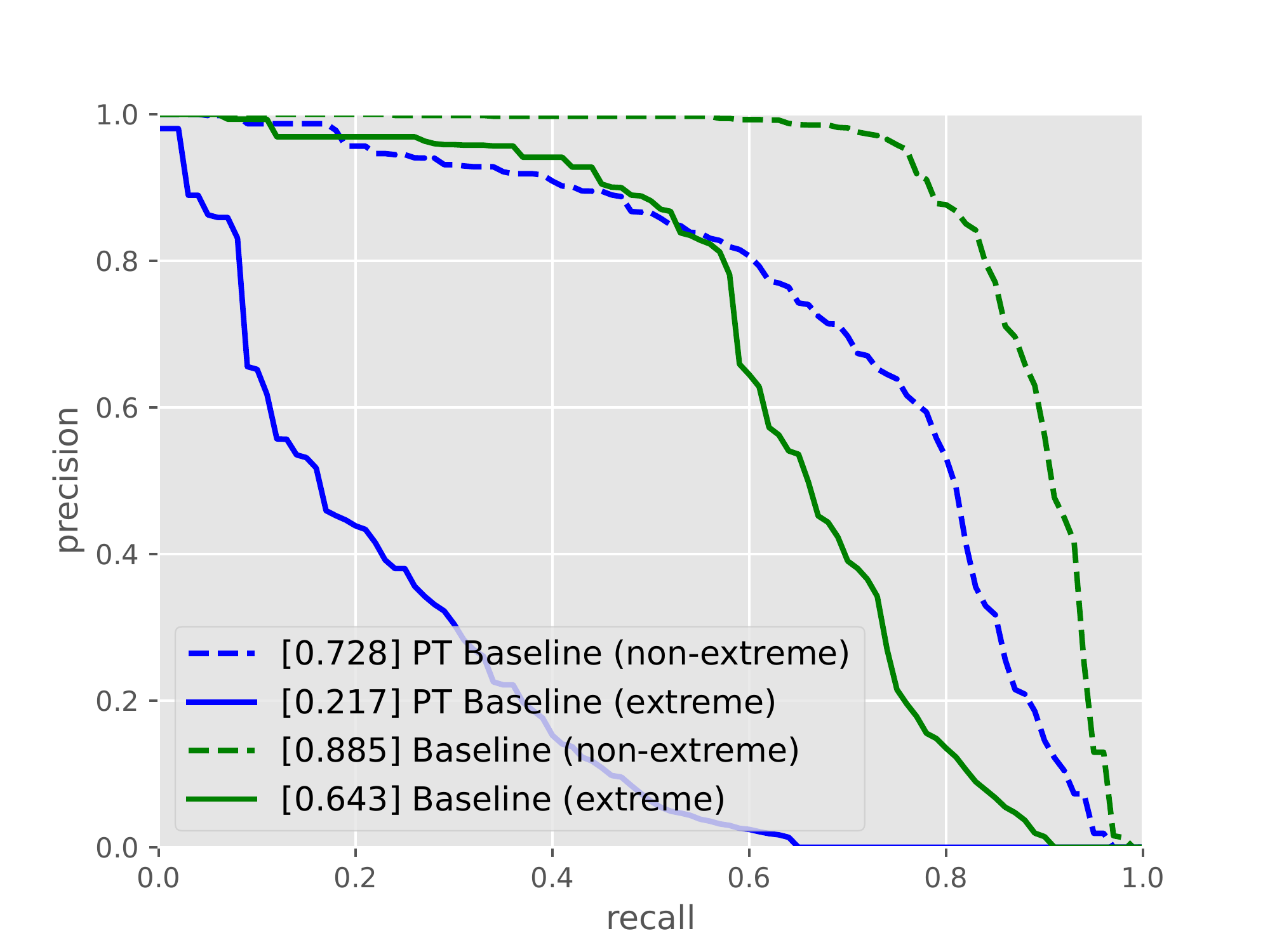}}
 & 
\includegraphics[width=2.75cm]{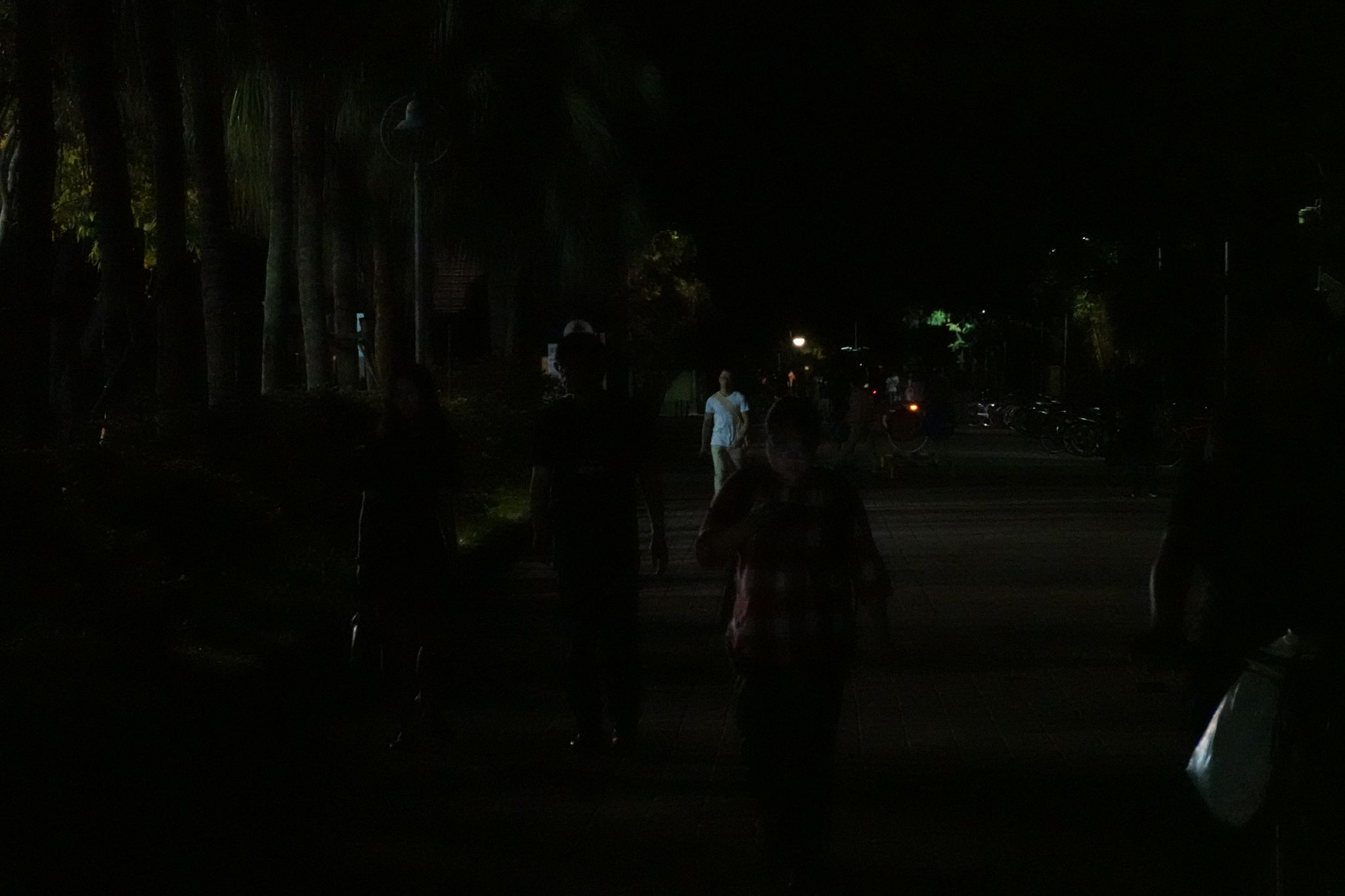} \includegraphics[width=2.75cm]{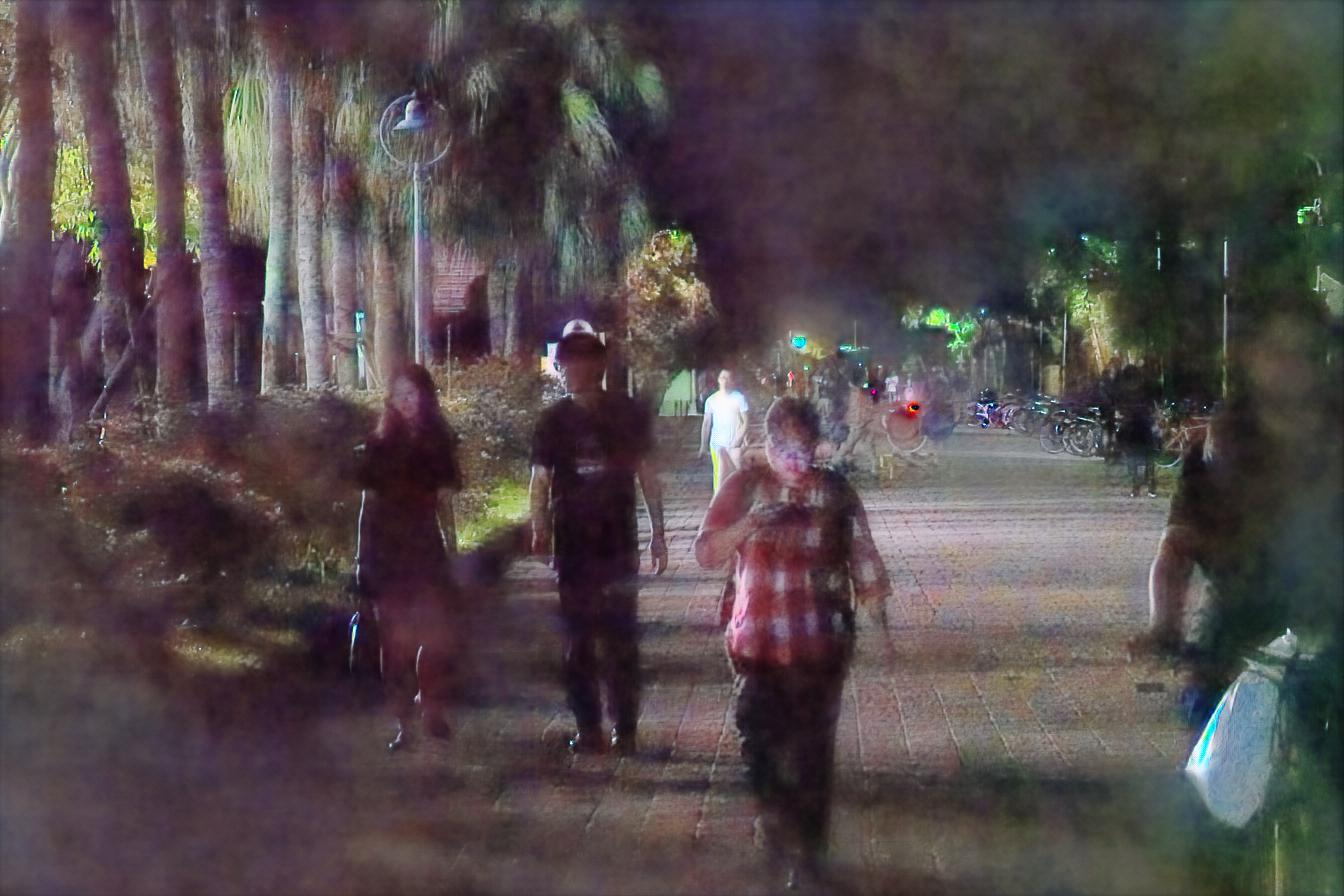} \\
 &
\includegraphics[width=2.75cm]{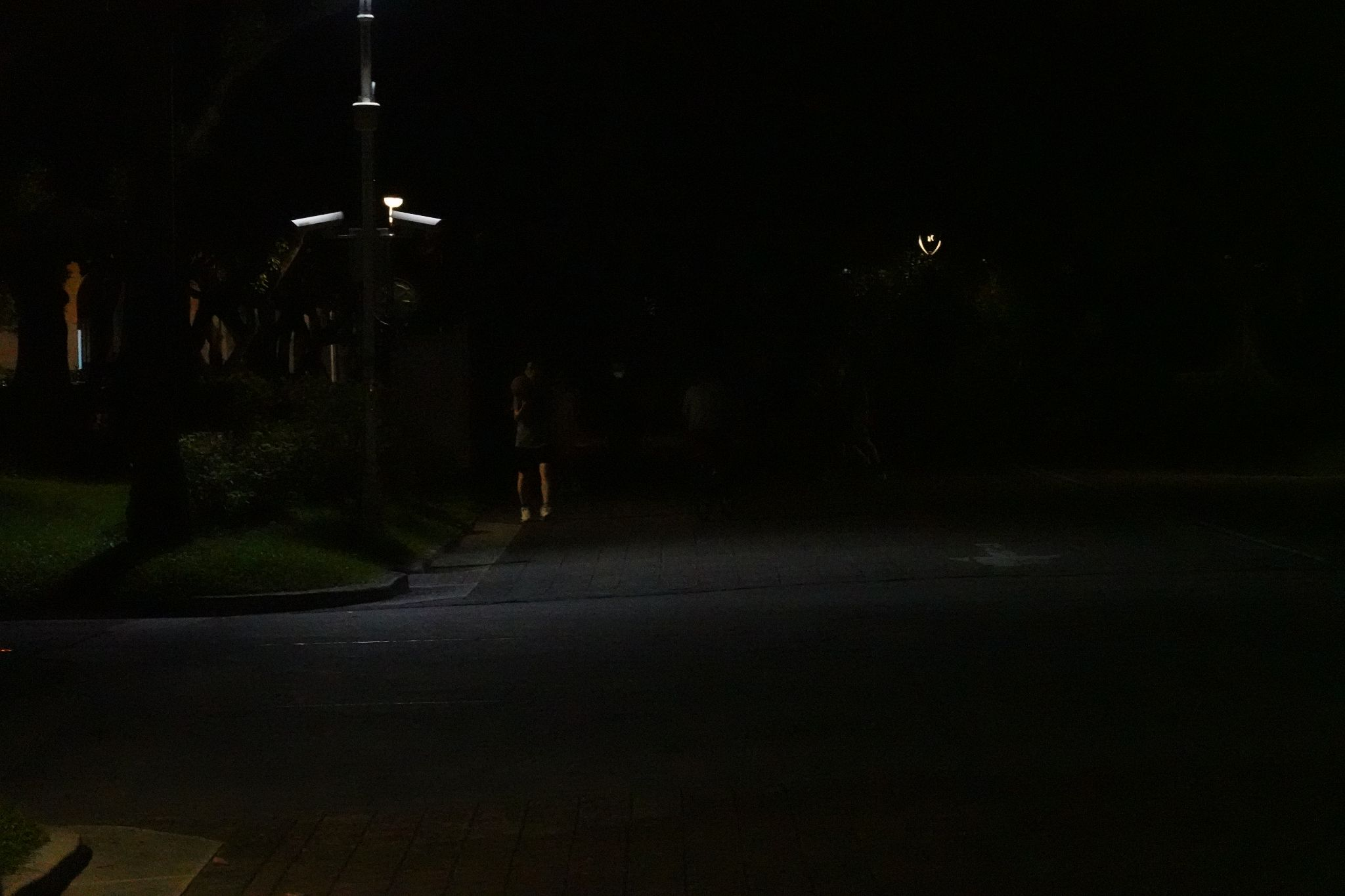} \includegraphics[width=2.75cm]{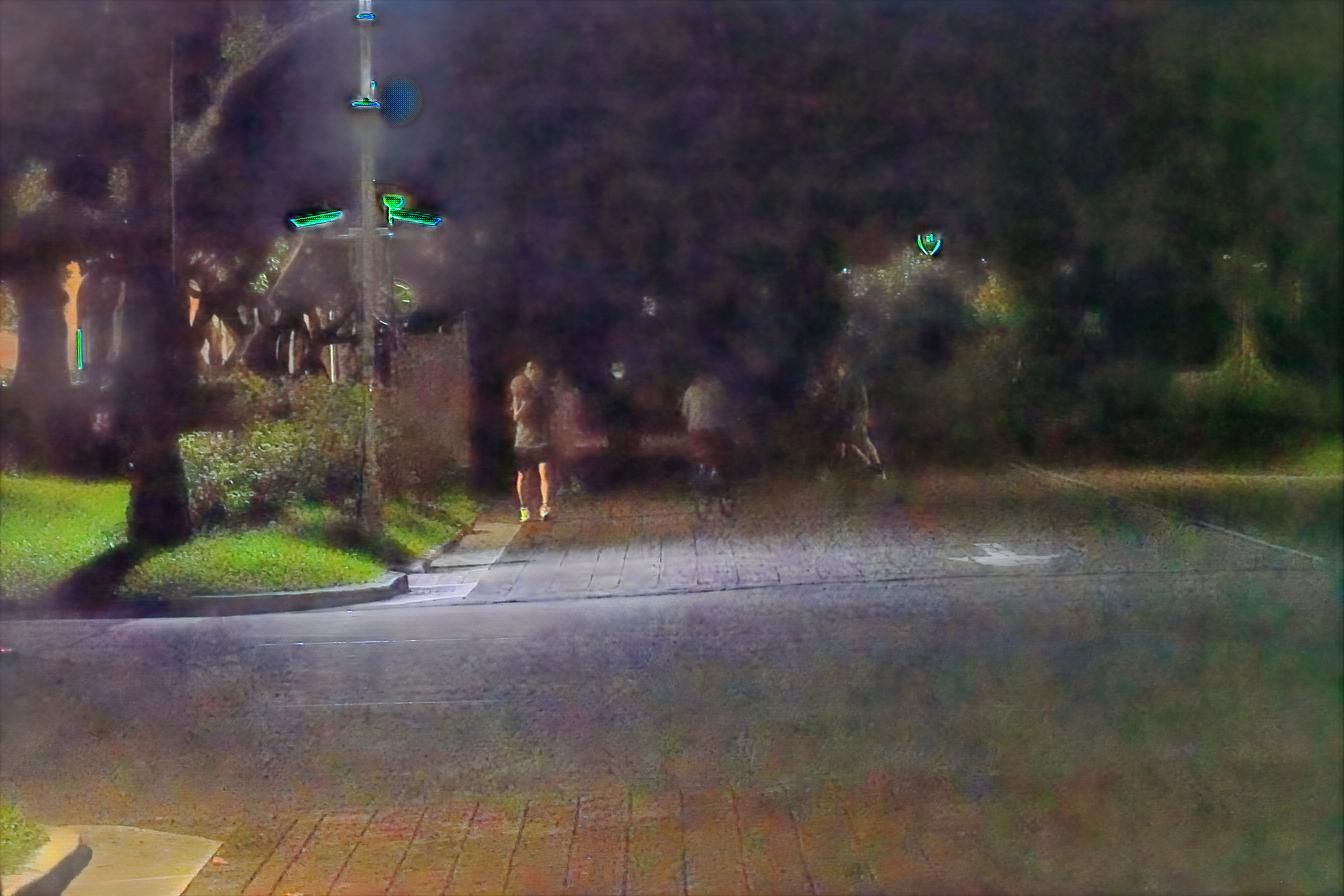}
\vspace{-0.2em} \\
(a) & \multicolumn{1}{c}{(b)} \vspace{-0.7em}
\end{tabular}
\end{center}
\caption{(a) Precision-Recall curves of the pre-trained baseline model (PT Baseline) and the baseline model fine-tuned on our dataset (Baseline) under \textit{extreme} and \textit{non-extreme} low-light conditions. Although training on an appropriate low-light dataset helps reducing the gap between \textit{extreme} and \textit{non-extreme} conditions, the gap is not entirely eliminated. This suggests that new methods targeting extreme low-light conditions specifically are required. \\ (b) Image input to our proposed detection-with-enhancement model (left) and enhanced intermediate representation (right).}
\label{fig:two}
\vspace{-1.5em}
\end{figure}

\vspace{-1.2em}\section{Introduction}
\vspace{-0.6em}\label{sec:intro}
In recent years, deep learning-based methods for object detection have achieved great success. Although the performance of current object detectors under normal conditions can be impressive, recent work \cite{Exdark,REG} shows that machine perception under low-light conditions turns out to be a complex problem that requires special attention from the researchers.

Fundamentally, difficulties in perception in low light stem from imaging problems caused by low photon count. Techniques in photography to address this are to: 1) gather more light by increasing the aperture size or extending the exposure time, and 2) increase the sensitivity by setting higher ISO values. However, these solutions lead to out-of-focus blur, motion blur, and amplification of noise, which prove difficult for machine perception.
 
An appropriate approach for learning-based frameworks is to increase the number of data presenting low-light conditions. Still, the issue of low-light is far from being solved, given that: 1) high-quality large-scale datasets containing a high proportion of low-light data are sparse and difficult to label, 2) \cite{Exdark} showed that even when trained on equal amounts of low-light and bright data ConvNets do not learn to normalize deep features with respect to the lighting conditions, \textit{i.e.,} low-light and bright features form two separate clusters of data, and thus require separate modeling.

We address these issues in three dimensions by: 1) releasing a novel dataset that can be used to study and develop models targeting images in low-light conditions, 2) analyzing the limitations of the baseline model on our dataset and gaining insight as to what exactly is difficult for machine cognition in low light, 3) developing a learning-based image enhancement module and novel augmentation techniques targeting low-light conditions.

Our first contribution is the Night Object Detection (NOD) dataset\footnote{Our dataset is publicly available at \href{https://github.com/igor-morawski/NOD}{https://github.com/igor-morawski/NOD}.}: a high-quality large-scale dataset captured in the wild under challenging low-light conditions, annotated with instances of \textit{people}, \textit{bicycles} and \textit{cars}. For a subset of the dataset, we provide instance-level annotations of lighting conditions in novel terms of \textit{extreme} and \textit{non-extreme} low-light conditions, for meaningful evaluation and benchmarking of the future object detection and image enhancement methods targeting low-light conditions.

Our second contribution is an analysis of how low-light conditions impact the performance of the baseline, where we show that the problem cannot be entirely solved by training on low-light data. We further link the lighting conditions to perceptual difficulty, and identify that there are \textit{non-extreme} low-light conditions that are moderately difficult for current object detectors, and \textit{extreme} low-light conditions that are very difficult for machine perception. Specifically, we define extreme low-light as a condition where most object edges and keypoints are not visible due to low illumination only, \textit{e.g.,} not due to occlusion.  As we show, training on an appropriate low-light dataset does not remove the performance gap between these two conditions (Fig. \ref{fig:two} (a)). Thus, new methods that target low-light conditions particularly are required.

Finally, we propose a method targeting low-light object detection that consists of an image enhancement module for intermediate image enhancement (Fig. \ref{fig:two} (b)) and two data augmentation methods. Accordingly, we present experimental results that show a consistent improvement over the baseline, including improvement under \textit{extreme} low-light conditions.
\vspace{-1.2em}\section{Related Work}
\vspace{-0.6em}\label{sec:related}
\textbf{Low-light image enhancement}. Learning-based solutions have been applied to numerous low-level vision tasks such as denoising, super-resolution, image enhancement, including low-light image enhancement.
Many works are inspired by Retinex theory that decomposes images into illumination and reflectance \cite{zhang2019kindling,zhang2021beyond,wang2019underexposed,wei2018deep}. However, Image Signal Processing (ISP) pipeline, used to produce JPEG images from raw data, breaks down under extreme low-light conditions, and thus, another line of work focuses on developing learning-based ISP pipelines \cite{SID,chen2019seeing,schwartz2018deepisp,ratnasingam2019deep}. Deep learning-based methods most often required paired training data and because of that most datasets are limited to static scenes \cite{SID} or synthetic data \cite{LORE2017650,lim2015robust,wang2019underexposed}. In contrast with these methods, we focus not on  improving perceptual quality, but on improving image representation for machine cognition in high-level tasks. 

\textbf{Low light in high-level vision tasks}. The closest to our work are \cite{Exdark,REG} and \cite{poor_visibility_benchmark}. Besides contributing a dataset of low-light images for image recognition and object detection, based on extensive investigation, \cite{Exdark} concluded that: 1) increasing amount of low-light data is necessary for improving low-light image cognition, 2) learned features extracted from the same object under good and poor lighting conditions belong to different data clusters. In our work, we continue investigation into machine cognition under low-light conditions, but we link the lighting conditions directly to perceptual difficulty rather than, \textit{e.g.,} light source as \cite{Exdark}. In comparison with their dataset, our dataset contains, on average, more annotated instances per object category, and the resolution of the images is higher. Another dataset under low-light conditions, DarkFace, a large-scale dataset for face detection under was released by  \cite{poor_visibility_benchmark}. In contrast with the DarkFace \cite{poor_visibility_benchmark} dataset, when an object was occluded in our dataset, we still annotated around the most probable boundary rather than around the visible part only. This is especially important in situations where a part of the object is not visible due to imaging difficulties. Finally, \cite{REG} proposed a detection-with-enhancement framework for low-light face detection based on the generation of multi-exposure images from a single image. Similarly, we propose to incorporate an image enhancement module. However, our image enhancement module produces single-exposure enhanced image representation. 

\vspace{-0.6em}\section{NOD: Night Object Detection Dataset}
\vspace{-0.6em}\label{sec:dataset}

\begin{table}[!h]
\vspace{-1.3em}
\footnotesize
\begin{center}
\begin{tabular}{c c c c c c c}
\toprule
Dataset & Camera & \# classes & \makecell{\# annotated \\ images} & \# instances & \makecell{\# unannotated \\ images} & \makecell{High- \\ Res.}\\
\hline
Sony & Sony RX100 VII & 3 & 3.2k & 18.7k & 0.9k & \checkmark \\
Nikon & Nikon D750 & 3 & 4.0k & 28.0k & 0 & \checkmark \\
\hline
NOD (ours) & Sony \& Nikon & 3 & 7.2k & 46.7k & 0.9k & \checkmark  \\ 
ExDark \cite{Exdark}& & 12 & 7.3k & 23.7k & 0 & \xmark \\
\toprule
\end{tabular}
\vspace{-.8em}
\end{center}
\caption{Basic statistics in the Night Object Detection (NOD) dataset. We provide high-quality bounding box annotation for \textit{people}, \textit{bicycles} and \textit{cars}.}
\label{tab:stats}
\vspace{-1.5em}
\end{table}

We present a high-quality large-scale dataset of outdoor images targeting low-light object detection. The dataset contains more than 7K images and 46K annotated objects (with bounding boxes) that belong to classes: \textit{person}, \textit{bicycle}, and \textit{car}. The photos were taken on the streets at evening hours, and thus all images present low-light conditions to a varying degree of severity. We used two DSLR cameras to capture the scenes: Sony RX100 VII and Nikon D750, and throughout the paper, we refer to the sets collected by these cameras as Sony and Nikon (data)set. We show the statistics of our dataset in Tab. \ref{tab:stats}.

All photos were shot handheld, and most of them were shot in Full Auto mode. Some of them shot in Shutter Priority mode, especially when there were fast moving objects (\textit{e.g.} cars) involved. Thus, the images in our dataset show all common culprits of low-light photography: motion blur, out-of-focus blur, and severe noise. 
To ensure the high quality of annotation under challenging conditions, we outsourced data labeling to a company that annotated instances on images enhanced by MBLLEN \cite{lv2018mbllen} in their original resolution. 

\vspace{-0.2em}
\begin{figure}[t]
\begin{center}
\begin{tabular}{ccccc}
\includegraphics[height=2cm]{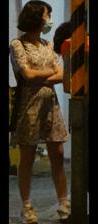} &
\includegraphics[height=2cm]{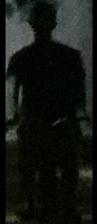} &
\includegraphics[height=2cm]{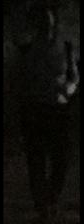} &
\includegraphics[height=2cm]{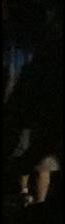} &
\includegraphics[height=2cm]{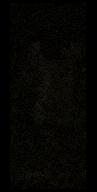} \vspace{-1mm} \\ 
\ding{117} & \begin{footnotesize}\ding{110}\end{footnotesize}& \ding{54} & \ding{116} & \ding{115}\vspace{-0.5em}
\end{tabular}
\begin{tabular}{cc}
\includegraphics[width=5.55cm]{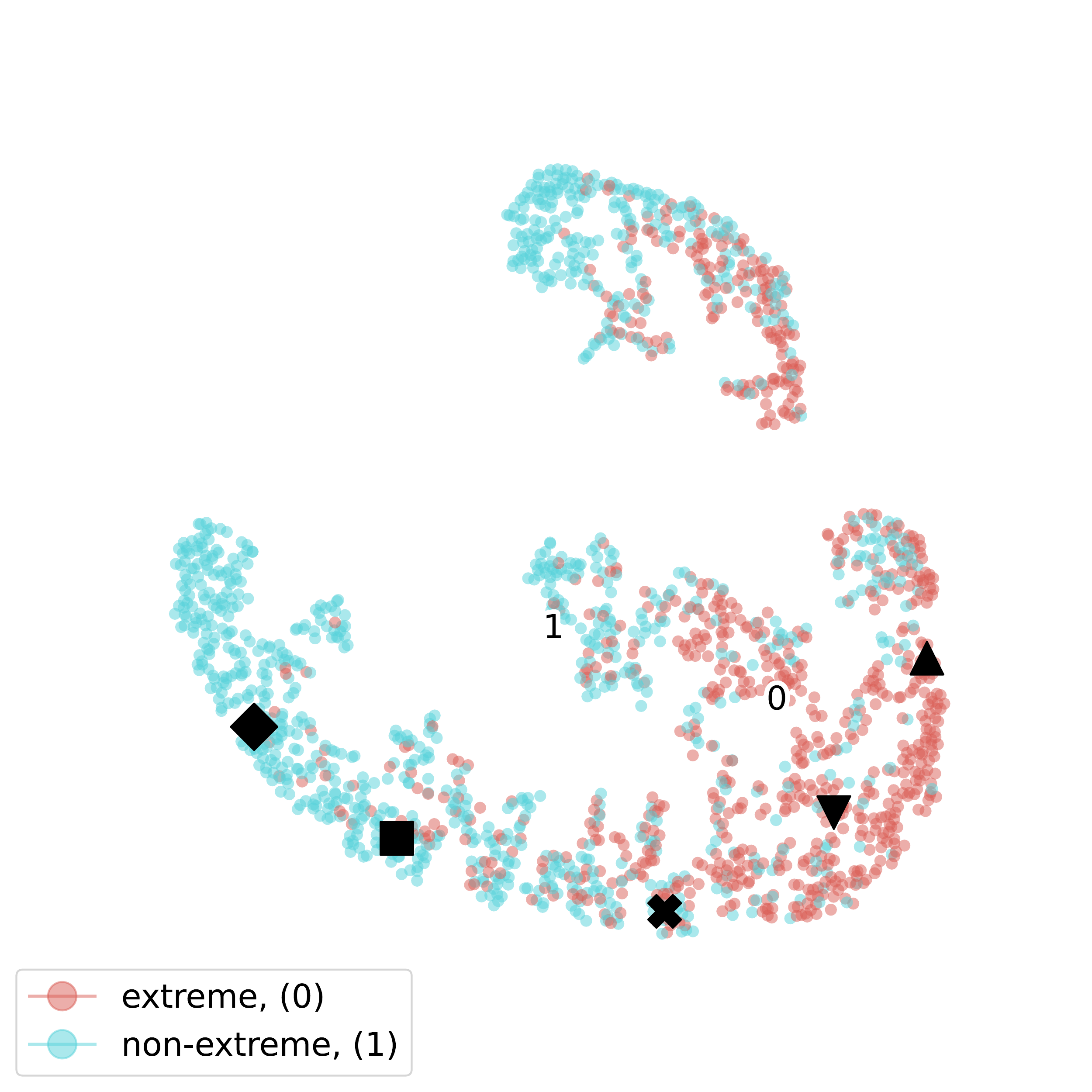} 
 & \includegraphics[width=5.55cm]{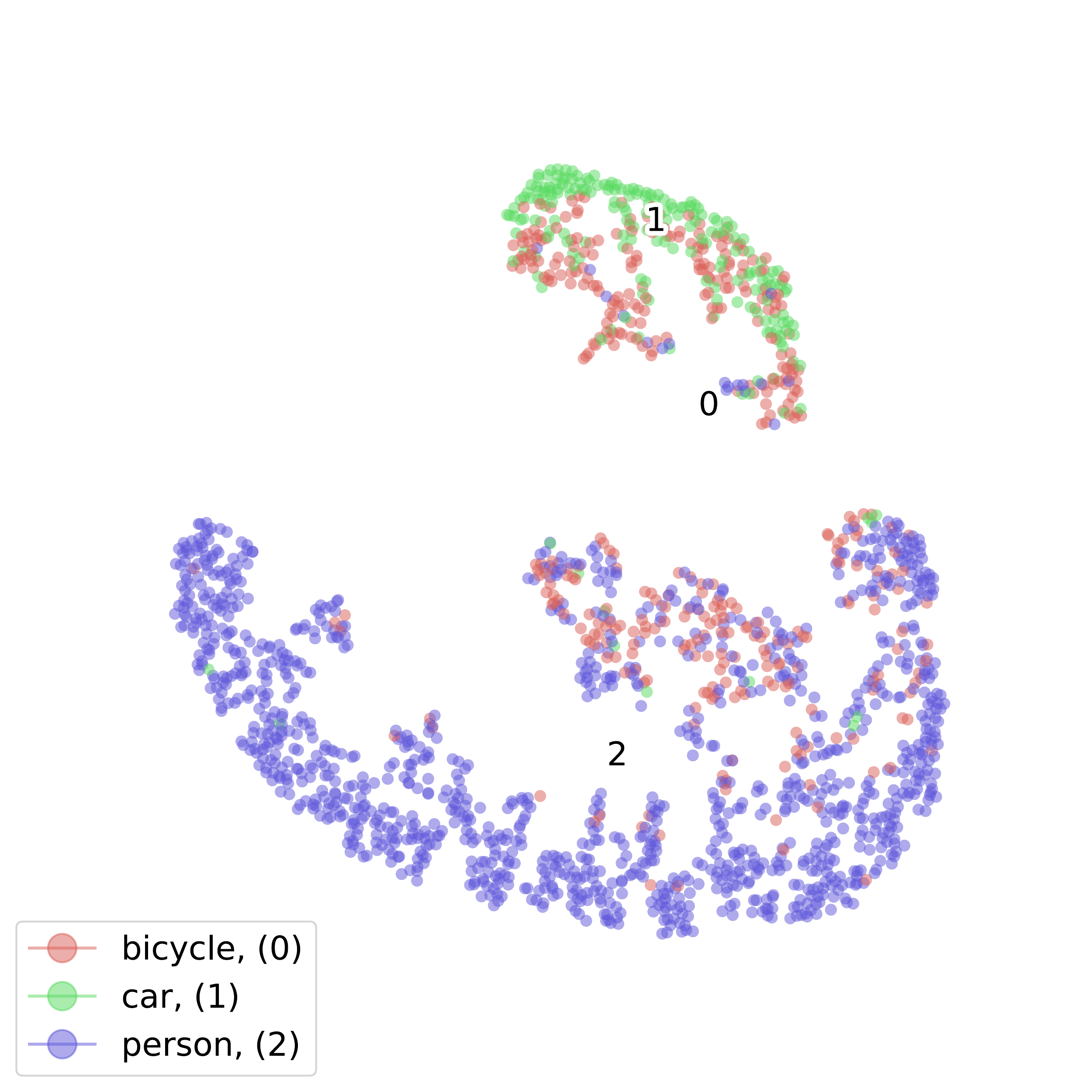} \vspace{-0.2em} \\
(a) & (b) \vspace{-0.7em}
\end{tabular}
\end{center}
\caption{t-SNE embeddings of the features extracted by the baseline model pre-trained on the COCO \cite{COCO} dataset. Rather than classifying by the lighting source, we directly link low-light conditions to perceptual difficulty. We define \textit{extreme} low-light conditions as conditions where most of the object edges are not visible due to poor illumination. To illustrate, \ding{117} and \begin{footnotesize}\ding{110}\end{footnotesize} belong to the same data cluster despite a large apparent difference, \textit{i.e.}, \ding{117} is well-illuminated and \begin{footnotesize}\ding{110}\end{footnotesize} is backlit.
At the same time, \ding{116} belongs to the \textit{extreme} conditions cluster, even though a part of the object (legs) is relatively well-illuminated and clearly visible.}
\label{fig:tsne}
\vspace{-1.5em}
\end{figure}

\begin{figure}
\begin{center}
\begin{tabular}{cc}
\includegraphics[width=5.55cm]{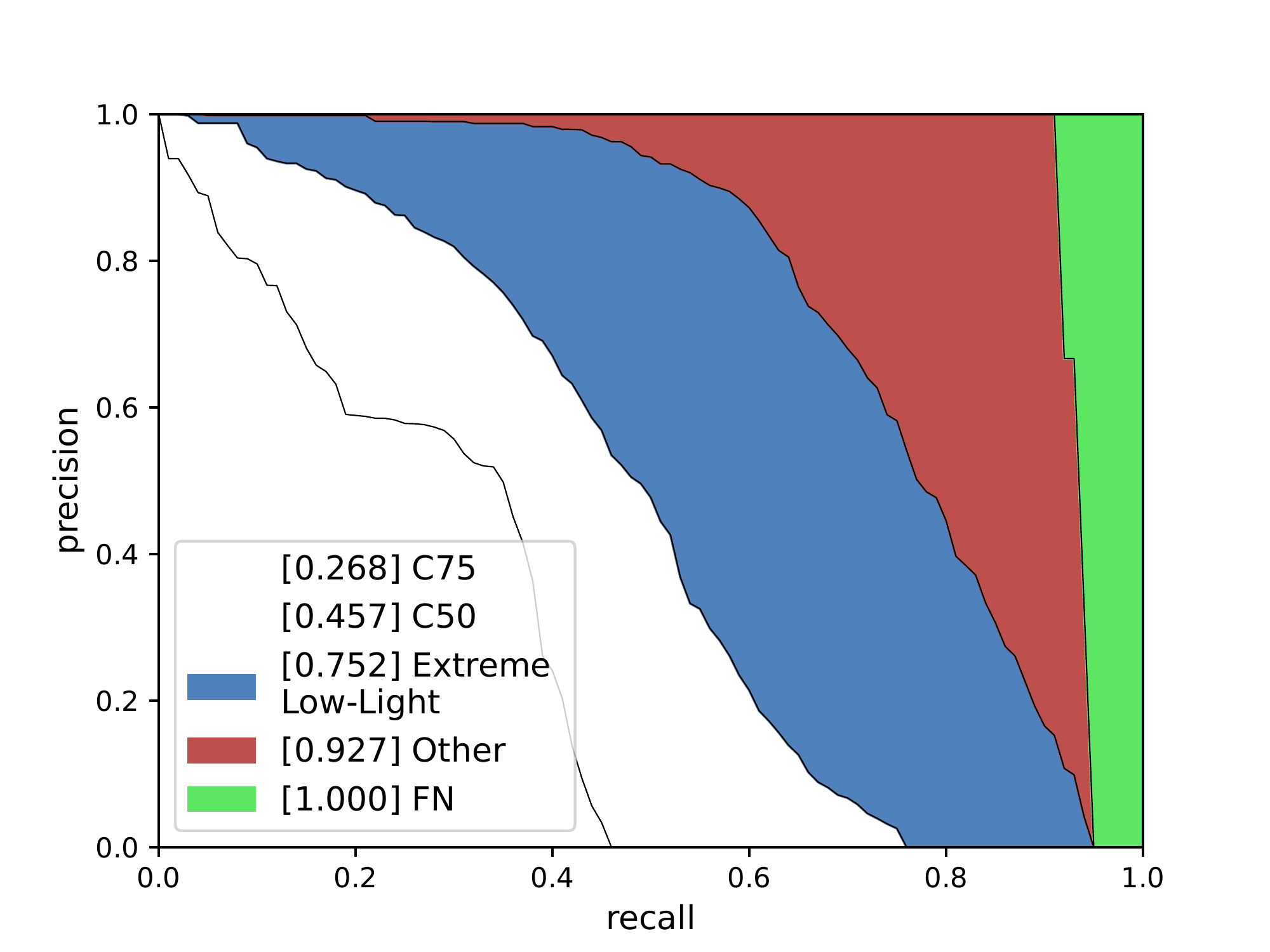} & \includegraphics[width=5.55cm]{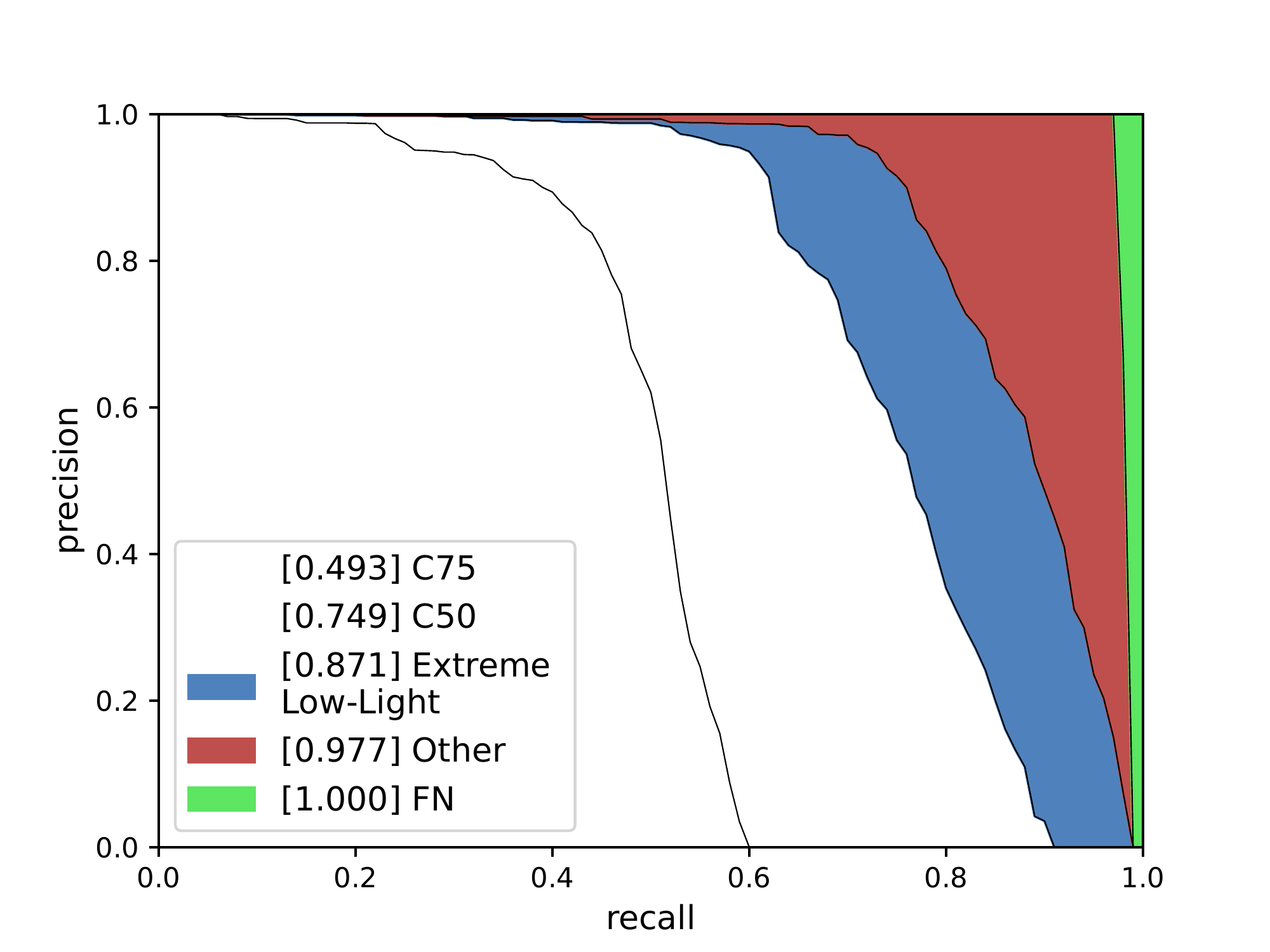}
\\ \vspace{-0.2em}
(a) & (b) \vspace{-0.7em}
\end{tabular}
\end{center}
\caption{(a) Precision-Recall curves of an off-the-shelf object detector on our dataset. Misdetections due to \textit{extreme} low-light conditions constitute a large part of errors in the baseline model. (b) PR curves of an object detector fine-tuned on our dataset. Although training the performance is significantly improved, errors due to \textit{extreme} low-light conditions still make up a substantial part of errors. \textit{C50} and \textit{C75} are PR curves at $IoU=0.5$ and $0.75$, respectively. \textit{Extreme Low-Light} is the PR curve after eliminating all misdetections that can be attributed to extreme low-light conditions. \textit{Other} is the PR curve after eliminating all misdetections that cannot be attributed to extreme low-light conditions.}
\label{fig:prs-anal}
\vspace{-1.5em}
\end{figure}

We target applications that focus on machine perception and end-task performance rather than applications such as enhancing perceptual quality. Therefore, in our dataset, we captured dynamic scenes in an uncontrolled environment that represent significant problems in photography under poor lighting. Over all, all images in our dataset present low-light conditions. However, the degree of severity of these conditions varies, and for more detailed evaluation, we provide instance-level annotation of lighting conditions for the test subset of the Sony set. We define extreme low-light as a condition where most object edges and keypoints are not visible due to low illumination only, \textit{e.g.,} not due to occlusion. Out of 1827 instances in this set, we manually labeled 810 as presenting extreme low-light conditions. Moreover, for this set, we also indicate if the object is truncated or strongly occluded. 

More details about the dataset, setup and annotation procedure as well as sample images and bounding box annotations can be found in the supplementary material.



\vspace{-1.2em}\section{Dataset and Baseline Analysis}\vspace{-0.6em}\label{sec:analysis}

To investigate whether differentiating between the extreme and non-extreme low-light conditions in this way is meaningful, we visualized t-SNE embedding of the features extracted by the backbone pre-trained on the COCO dataset \cite{COCO} with less than 0.23\% of images presenting low-light conditions \cite{Exdark}. We show the result colored using two mappings: by object class and by lighting conditions, in Fig. \ref{fig:tsne} (a) and (b), respectively. Indeed, the features extracted from regions presenting \textit{extreme} and \textit{non-extreme} low-light conditions belong to different data clusters. The observation that ConvNets do not normalize lighting conditions is in line with the observation made by \cite{Exdark} for image recognition networks. However, our classification of lighting types is by perceptual difficulty rather than by, \textit{e.g.}, the light source as in \cite{Exdark}. Moreover, we observe that there is no sharp boundary between these conditions, which seems to follow the intuition that lighting conditions are a spectrum from bright (easy for perception) through low-light (moderately difficult for perception) to extreme low-light (difficult for perception). Similarly, we visualized t-SNE embeddings for all the models in our paper and found out that these findings hold for all of them. 

Next, we analyzed the performance of an off-the-shelf detector on challenging low-light data. To this end, similarly as for visualizing t-SNE feature embeddings, we used the baseline model trained on the COCO dataset \cite{COCO}. We evaluated the performance on the test set of Sony, and used the lighting conditions annotations to investigate the impact of extreme low-lighting conditions on the detector. We show the results in Fig. \ref{fig:prs-anal} (a), where we observe that errors due to extreme low-light conditions constitute a large part of errors of the off-the-shelf detector. 
Similarly, we analyzed the performance of the same model fine-tuned on our low-light dataset, shown in Fig. \ref{fig:prs-anal} (b). Although the detector trained on low-light data performs much better under low-light conditions, the proportion of errors due to extreme low-light remains significant, and there is a large room for improvements in this aspect.

In order to verify that the performance gap between the pre-trained and fine-tuned baseline model is, indeed, due to the lack of low-light data rather than the distribution mismatch only, we compared the Precision-Recall curves separately under \textit{extreme} and \textit{non-extreme} low-light conditions. The results are shown in Fig. \ref{fig:two} (a). In comparison with the baseline fine-tuned on our dataset, the performance of the pre-trained baseline model under \textit{extreme} low-light conditions is disproportionately lower with respect to the performance under \textit{non-extreme} low-light conditions. In other words, training on an appropriate low-light dataset, helps to reduce the gap between the performance under extreme and non-extreme lighting conditions. However, despite the large amounts of extreme low-light data in training, the gap is not entirely reduced, which suggests that special attention from the researchers is required to solve the problem of low-light conditions in high-level tasks.

\vspace{-0.6em}\section{Proposed Method}
\begin{figure}[t]
\begin{center}
\includegraphics[width=10cm]{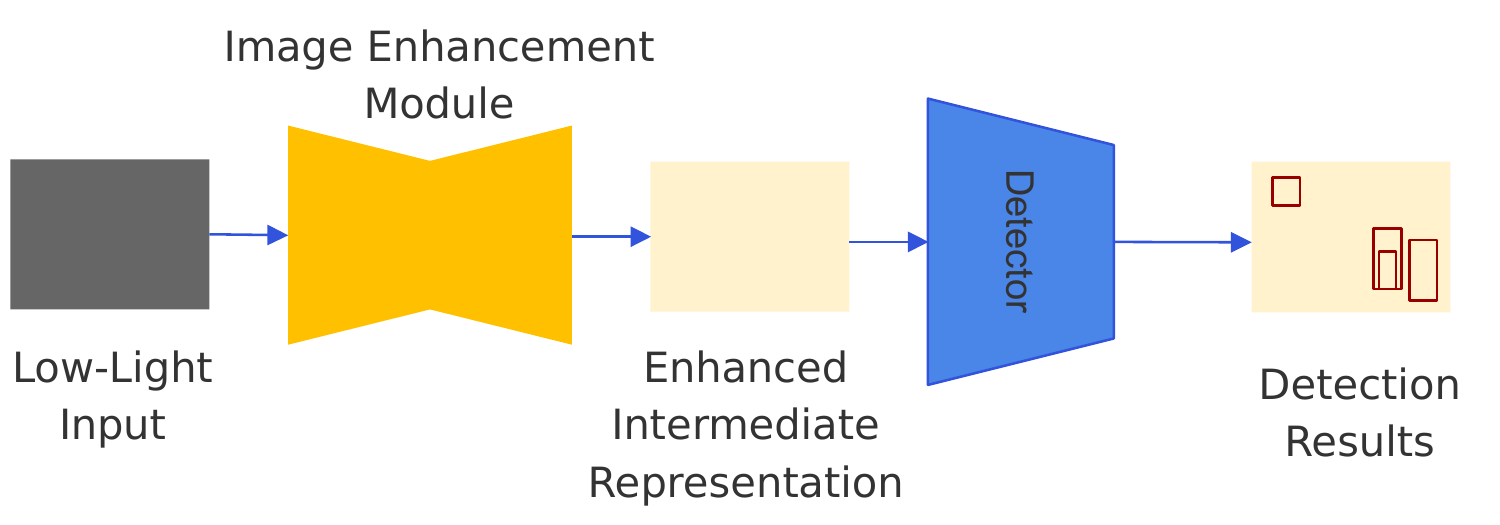}
\vspace{-1.9em}
\end{center}
\caption{The proposed method consists of an image enhancement module trained under the guidance of the object detector that produces enhanced intermediate representation optimal for machine cognition. }
\vspace{-1.5em}
\label{fig:method}
\end{figure}

\vspace{-0.6em}\label{sec:method}
Inspired by the observation that ISP pipelines are not designed to work under extreme low-light conditions, we introduce an image enhancement module that will compensate for the errors of the ISP pipeline as shown in Fig. \ref{fig:method}. The image enhancement module is to compensate for extreme low-light conditions and is trained jointly with the object detector to learn image representation optimal for machine cognition rather than for the human visual system. 

In our exploratory study, we have experimented with image-to-image fully-convolutional networks and image-to-parameter networks. In the end, we have selected U-Net \cite{unet} as an effective architecture for this task. In contrast with image-to-parameters networks, such architecture is capable of both performing intensity adjustment as well as denoising. We have also found out that training U-Net from scratch jointly with the object detector initialized from a pre-trained checkpoint leads to suboptimal results. Therefore, we propose a simple but effective pre-training procedure that takes advantage of large amounts of bright images easily available for training. 

\vspace{-1.2em}\subsection{Pre-Training Image Enhancement Module}\vspace{-0.6em}
We propose a pseudo-supervised pre-training procedure that leverages the abundant amount of data collected under normal conditions. Two observations inspired our approach: under extreme low-light conditions, 1) low SNR is one of the critical factors limiting image quality, and 2) because of the relatively low bit-depth of JPEG data, naively applying brightness and contrast adjustment leads to visual artifacts that are similar to the posterization effect.

We formulate the pre-training as an image restoration task, with original bright images as target images. We extract random patches from images and corrupt well-lit data by first reducing the number of gray levels to $k$ (\textit{i.e.}, we use $k$ levels to represent 256 gray levels, using \textit{e.g.} uniform color quantization), and then adding shot noise on top of posterized image patches. The number of gray levels and noise parameters can be used to control the severity of image corruption. Examples of the corrupted images can be found in the supplementary material.


We train the image enhancement module from scratch, and measure the distance between the original image $I$ and the reconstructed image $\hat{I}$ using pixel-wise MSE and the Structure Similarity (SSIM) index \cite{ssim}. Moreover, we use VGG loss \cite{vggloss}, to encourage the network to focus on high-level image features rather than low-level statistics, crucial to the object detection task. The total loss is formulated as below:\vspace{-.7em}
\begin{equation}
\pazocal{L} = MSE(I,\hat{I}) + \lambda_1 SSIM(I,\hat{I}) + \lambda_2 VGG(I,\hat{I})
\vspace{-.9em}
\end{equation}
where $\lambda_1$, $\lambda_2$ are hyper-parameters.

\begin{figure}
\begin{center}
\begin{tabular}{cc}
\includegraphics[width=5.55cm]{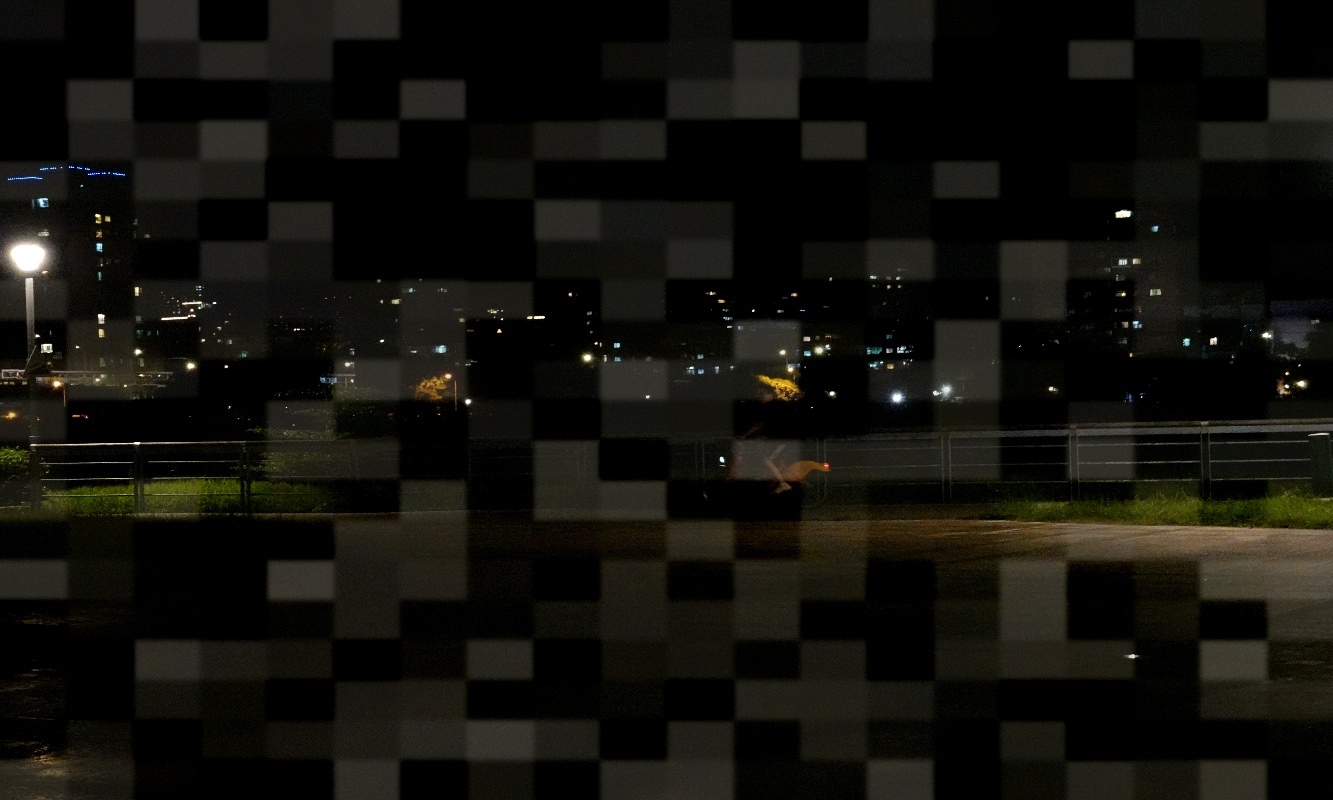} & 
\includegraphics[width=5.07cm]{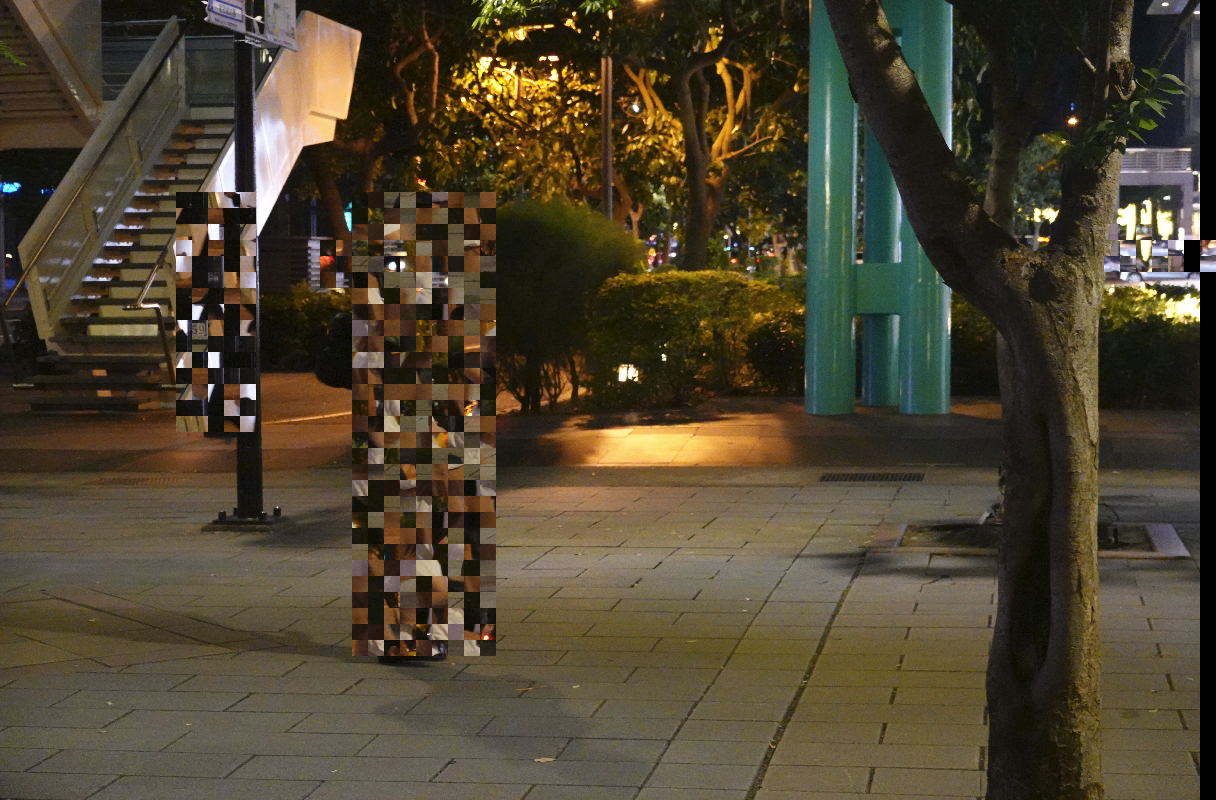} \vspace{-0.2em} \\
(a) & (b) \vspace{-0.7em}
\end{tabular}
\end{center}
\caption{Proposed augmentation methods: (a) patch-wise light augmentation to reduce the spatial redundancy, (b) block shuffle augmentation to encourage the detector to look at the object's context. Best viewed on display.}
\vspace{-1.5em}
\label{fig:aug}
\end{figure}

\vspace{-1.2em}\subsection{Patch-Wise Light Augmentation}\vspace{-0.6em}
In night scenes, local illumination changes are common due to the presence of local light sources and shadows cast by objects. Under- and over-exposed regions are another example of local lighting variations common in low-light photography. Ideally, a good image enhancement module should learn to remove these variations. To facilitate the learning of compensating these local variations, we propose to remove the spatial redundancy in the images by randomly adjusting brightness and contrast in every patch of the input image. We adjust brightness and contrast using the formula: \vspace{-.7em}
\begin{equation}
I'(x,y)=\alpha(x,y)(I(x,y) + \delta(x,y) ),
\vspace{-.9em}
\end{equation}
where $I(x,y)$ is the original intensity, $\alpha$ controls contrast adjustment, and $\delta$ controls brightness adjustment. For each patch, $\alpha$ and $\delta$ are sampled randomly from a range of possible values. 
An example of patch-wise light augmentation is shown in Fig. \ref{fig:aug} (a).

\vspace{-1.2em}\subsection{Block Shuffle Augmentation}\vspace{-0.6em}
Under extreme low-light conditions, object features are severely affected by lighting variations, \textit{e.g.,} edges or object keypoints are not visible. In such a case, humans may approach the problem by paying closer attention to the object's context rather than only the object itself. In order to encourage the object detector to do the same, we propose to destroy the spatial correlation of the features by "scrambling" each object region. For each object region, if it's selected for augmentation with probability $p$, we divide the region into blocks sized $B\times B$ and randomly permute them. An example of this augmentation is shown in Fig. \ref{fig:aug} (b). In this way, the detector is forced to look outside the scrambled region to look for more clues about the object. We hypothesize that this augmentation might also boost object detection performance by augmenting features outside their normal context. 
\vspace{-0.6em}\section{Experimental Results and Discussion}
\textbf{Implementation Details}.
We implement all models with Open MMLab Detection Toolbox \cite{mmdetection} on 2 Tesla-V100 32GB GPUs with SyncBN. We use SGD optimizer, apply a batch size of 8, and set the learning rate to $\expnumber{1}{-4}$. As for the U-Net \cite{unet}, we replace ReLU activatations with Mish \cite{mish} and add Batch Norm layers before every activation layer. We  extract random patches from the COCO dataset \cite{COCO}, and corrupt them by applying posterization, by reducing from 256 to $k\in [2,8]$ gray levels, and adding shot noise. We reduced gray levels by uniform color quantization, \textit{i.e.}, we divided each axis of the color space into equal sized segments. We use Adam \cite{adam}, apply a batch size of 64, aset the learning rate to $\expnumber{1}{-4}$, and train using two Teslas K80 12GB. As for the patch-wise augmentation, we set $\alpha$, $\delta$ limits to $[-0.3,0.3]$, and vary patch size from 4\% to 20\% of the image size during training. As for the block shuffle augmentation we set block size to $16\times 16$. More details can be found in the supplementary material.

\vspace{-1.2em}\subsection{Experiments \& Discussion}\vspace{-0.6em}
We first present the ablation study to show the impact of the proposed improvements on the performance. The effectiveness of the proposed pre-training method is shown in Tab. \ref{tab:pretraining}, the ablation study of the enhancement module is shown in Tab. \ref{tab:rebuttal-camera-abl}, and the effectiveness of the proposed image enhancement module and augmentation methods are shown in Tab. \ref{tab:ablation}. We also present images enhanced by the proposed image enhancement module in Fig. \ref{fig:two} (b) and in the supplementary materials.

\begin{table}[ht]\vspace{-0.7em}
\footnotesize
\begin{minipage}[b]{0.45\linewidth}
\centering
\begin{tabular}{c c c c}
\toprule
Pre-training & $AP_{50}$ & $AP_{75}$ & $AP$ \\
\hline
 \xmark & 64.9\% & 41.2\% & 40.3\% \\
 \checkmark & \textbf{74.0\%} & \textbf{48.0\%} & \textbf{47.1\%} \\
\toprule 
\end{tabular}
\caption{Effectiveness of our pre-training method. }
\label{tab:pretraining}
\end{minipage}
\hspace{0.5cm}
\begin{minipage}[b]{0.45\linewidth}
\centering
\begin{tabular}{c c c c c}
\toprule
Test & Enh. &   &   &   \\
dataset & module & $AP_{50}$ & $AP_{75}$ & $AP$ \\
\hline
\multirow{2}{*}{Sony} & \xmark & 72.8 \% & 47.6 \% & 45.9\% \\
& Nikon & \textbf{73.3\%} & \textbf{47.8\%} & \textbf{46.3\%} \\
\toprule 
\end{tabular}
\caption{Ablation study. We train the enhancement module on the Nikon subset, freeze the weights, and append it to the detector trained on the Sony subset.}
\label{tab:rebuttal-camera-abl}
\end{minipage}
\vspace{-1.4em}
\end{table}


\begin{table}[h]
\footnotesize
\begin{center}
\begin{tabular}{c c c c c c c c c c}
\toprule
 & \multicolumn{2}{c}{Augmentation} & \multicolumn{3}{c}{Sony} & \multicolumn{2}{c}{Nikon} \\
Backbone & Light & \makecell{Shuffle} & $AP_{50}$ & $AP_{75}$ & $AP$ & $AP_{50}$ & $AP_{75}$ & $AP$ \\
\hline
R(esNet)50& & & 
72.8\% &
47.6\% &
45.9\% & 69.7\% & 43.7\% & 44.0\%\\ 
\makecell{U-Net + R50} & 
 &
 &
74.0\% & 
48.0\% &
47.1\% &
70.8\% &
43.3\% &
44.4\% \\ 
\makecell{U-Net + R50}& 
 \checkmark & &
 73.9\% &
48.8\% & 
46.9\% & 
70.9\% & 
\textbf{45.0\%} & 
44.9\% \\ 
\makecell{U-Net + R50} & & \checkmark &
\textbf{74.3\%} &
49.0\% &
\textbf{47.5\%} &
71.2\% & 
44.7\% & 
44.8\% \\ 
\makecell{U-Net + R50} & \checkmark & \checkmark &
74.1\% &
\textbf{49.1\%} &
\textbf{47.5\%} &
\textbf{71.6\%} &
44.9\% &
\textbf{45.4\% } \\ 
\toprule
\end{tabular}
\vspace{-1.em}
\end{center}
\caption{Ablation study of the proposed method. We use RetinaNet \cite{lin2017focal} as the object detection framework and extend it by an image enhancement module if indicated by U-Net.}
\label{tab:ablation}
\vspace{-1.5em}
\end{table}

\begin{table}[t]
\footnotesize
\begin{center}
\begin{tabular}{c c c c c}
\toprule
Dataset & Proposed Method & $AP_{50}$ & $AP_{75}$ & $AP$ \\
\hline
\multirow{2}{*}{Sony} 
 & \xmark & 72.8\% & 47.6\% & 45.9\% \\
 & \checkmark &\textbf{ 74.1\%} & \textbf{49.1\%} &\textbf{ 47.5\%} \\
 \hline
\multirow{2}{*}{Nikon} 
 & \xmark & 69.7\% & 43.7\% & 44.0\% \\
 & \checkmark &\textbf{ 71.6\% }& \textbf{44.9\%} &\textbf{ 45.4\%} \\
 \hline
\multirow{2}{*}{NOD (Sony+Nikon)} 
 & \xmark & 73.1\% & 47.0\% & 46.3\% \\
 & \checkmark & \textbf{74.4\%} &\textbf{ 48.3\%} & \textbf{47.6\%} \\
 \hline
\multirow{2}{*}{ExDark \cite{Exdark}} 
 & \xmark & 78.3\% & 52.9\% & 48.7\% \\
 & \checkmark & \textbf{79.1\%} &\textbf{53.6\%} & \textbf{49.4\%} \\
 \hline
\toprule
\end{tabular}
\vspace{-1.em}
\end{center}
\caption{Results of the proposed method, including proposed augmentation methods, on subsets of our dataset (Sony, Nikon), our dataset (Sony+Nikon) and ExDark \cite{Exdark}. In all experiments, we use RetinaNet \cite{lin2017focal} as the object detection framework.}
\label{tab:overall}
\vspace{-1.5em}
\end{table}

We observe that the proposed image enhancement module improves the overall average precision, although it is more effective for the Sony dataset than for Nikon. We also observe that the augmentation methods have different impact on the performance depending on the dataset. For the Sony dataset, block shuffle augmentation is more effective, and for the Nikon dataset patch-wise light augmentation is more effective. Overall, the proposed method is effective in improving performance under low-light conditions. Next, we collected the overall results of the proposed method in Tab. \ref{tab:overall}. The proposed method shows a consistent improvement over the baseline model on our dataset as well as the ExDark dataset \cite{Exdark}.

\begin{figure*}[!h]
\begin{center}
\begin{tabular}{cc}
\includegraphics[width=5.55cm]{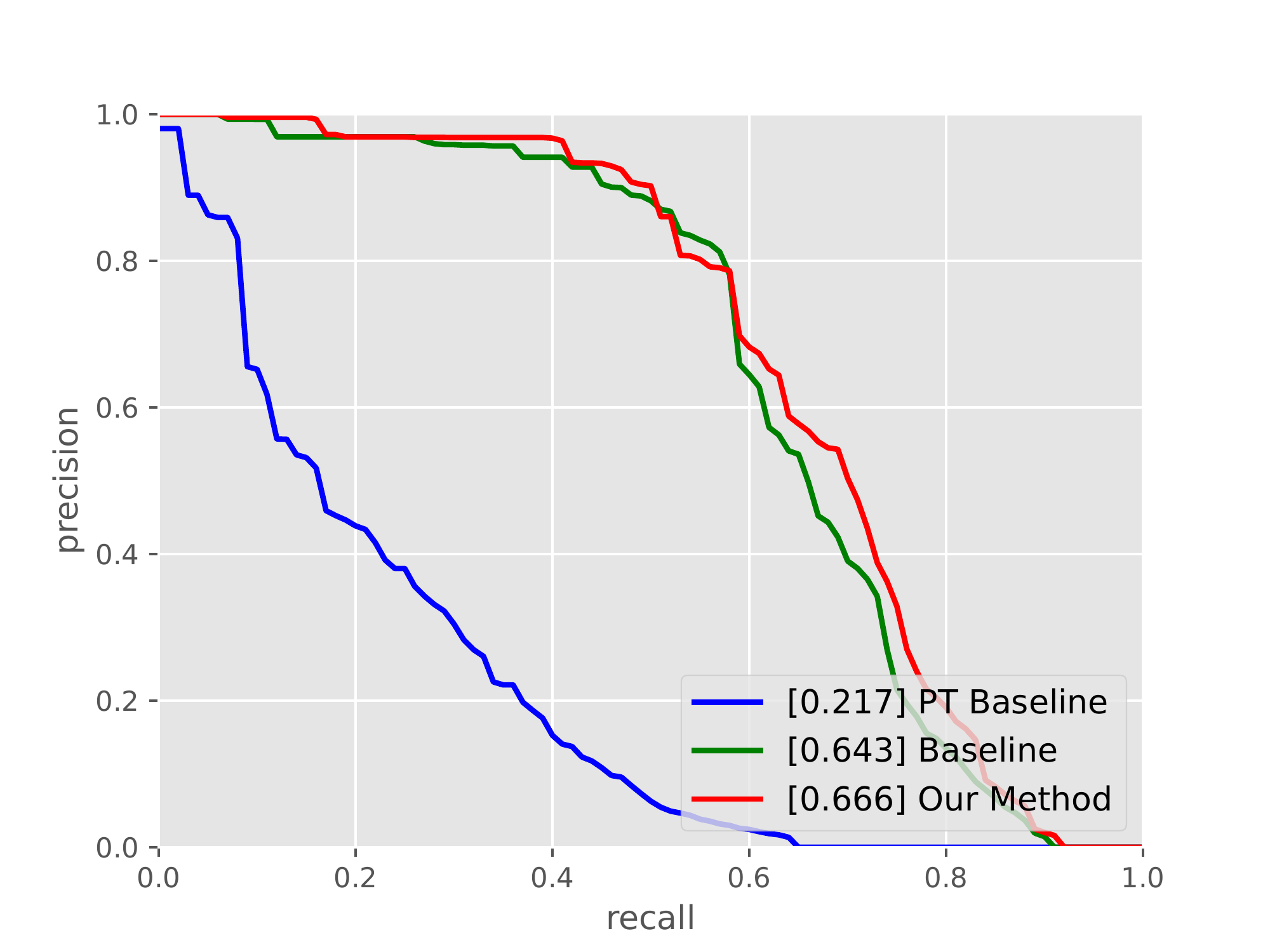} 
& \includegraphics[width=5.55cm]{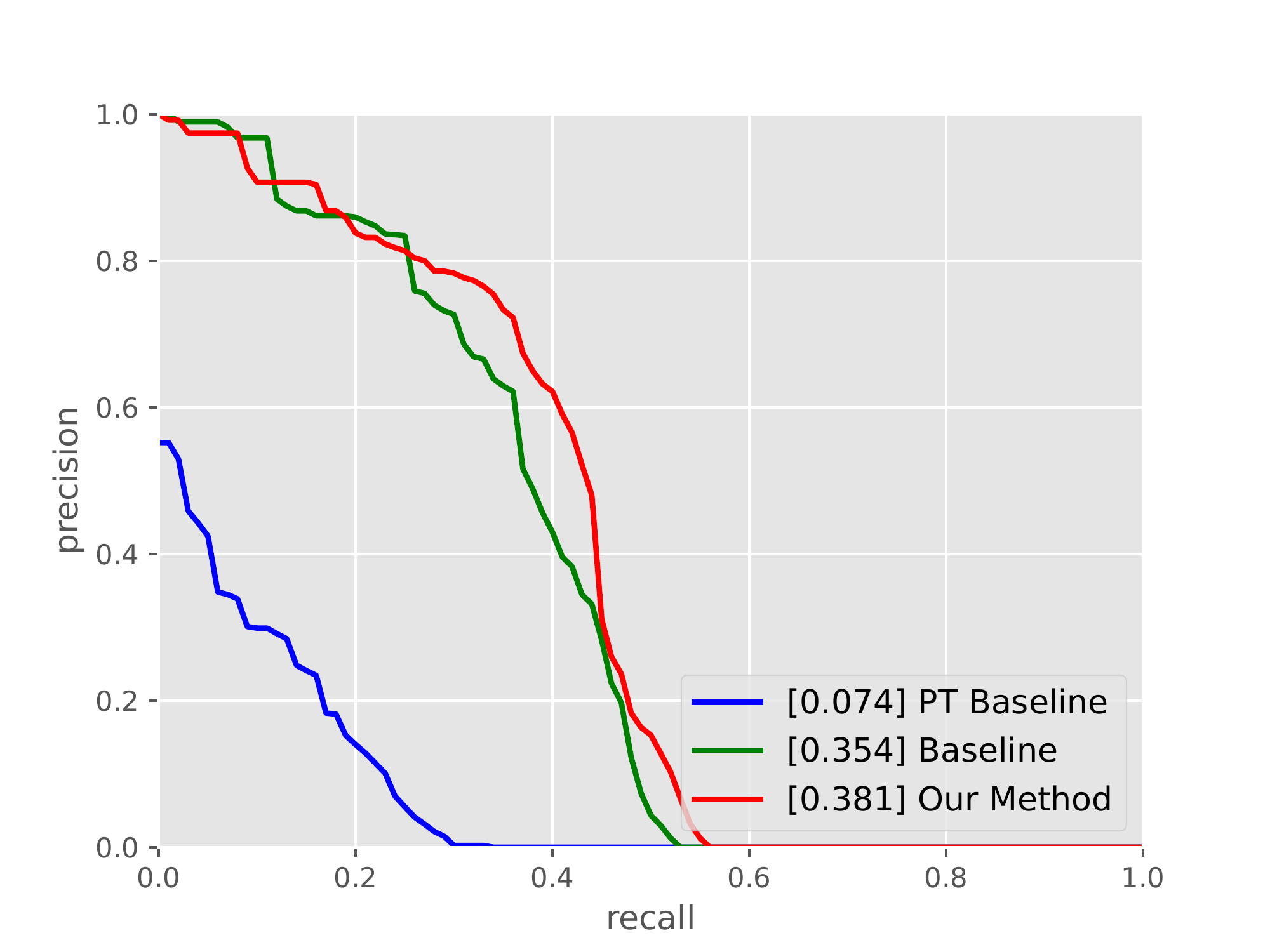} 
\vspace{-0.2em}  \\
(a) & (b)  \vspace{-0.7em}
\end{tabular}
\end{center}
\caption{\label{fig:three}
Precision-Recall curves under extreme low-light conditions, evaluated at (a) $IoU=0.5$ and (b) $IoU=0.75$. 
Under extreme-low light conditions, our method leads to more precise detection at the same recall level. Strongly occluded and truncated annotations were excluded from this evaluation. 
}
\vspace{-1.em}
\end{figure*}

However, as we showed in Subsection \ref{sec:analysis}, low light is a spectrum from \textit{non-extreme} conditions that are relatively easy for human and machine perception to \textit{extreme} conditions that are difficult for perception. To validate that the proposed method eliminates errors due to extreme low-light conditions rather than eliminating some other unrelated error, we evaluated the model on the instances in the Sony dataset presenting extreme low-light conditions only. The resulting Precision-Recall curves are shown in Fig. \ref{fig:three}. Our method has higher APs at $IoU=0.5$ and $0.75$, and generally shows a higher precision at the same recall level.

\vspace{-1.2em}


\begin{table}[!h]
\footnotesize
\begin{center}
\begin{tabular}{c c c c c c c c c c}
\toprule Enhancement & Requires & Learning- & Model & Model
  & \multicolumn{2}{c}{Extreme} & \multicolumn{2}{c}{Non-Extreme} \\
method& bright gt. & -based & GFLOPS  & parameters & $AP_{50}$ & $AP_{75}$ & $AP_{50}$ & $AP_{75}$\\
\hline
Ours & \xmark &  \checkmark & 12.1 & 8.0M & \textbf{63.7\%} & 35.2\% & 87.7\% & \textbf{71.2}\% \\
KinD++ \cite{zhang2021beyond} & \checkmark &  \checkmark & 7.9 & 8.3M & 63.5\% & \textbf{35.3\%} & \textbf{88.5\%} & 70.6\% \\
Zero-DCE \cite{Zero-DCE} & \xmark &   \checkmark & 5.2  & 79K & 61.3\% & 31.1\% & 87.3\% & 69.3\% \\
LIME \cite{LIME} & \xmark & \xmark & - & - & 60.8\% & 32.7\% & 87.4\% & 69.3\% \\ 
Hist. equal. & \xmark & \xmark & - & - & 60.1\% & 30.2\% & 86.2\% & 67.5\% \\ 
\toprule
\end{tabular}
\vspace{-1.5em}
\end{center}
\caption{Comparison to the related work in low-light image enhancement (training and testing on the Sony subset). 
The computational complexities are given for an input of size $256\times 256 \times 3$. The computational complexity of the baseline model (RetinaNet) is 13.1 GFLOPS. Strongly occluded and truncated annotations were excluded from this evaluation. }
\label{tab:rebuttal-SOTA}
\vspace{-1.5em}
\end{table}

We also show a comparison of our proposed enhancement module with other low-light enhancement modules after fine-tuning the baseline model in Tab. \ref{tab:rebuttal-SOTA}. Although the performance using KinD++ \cite{zhang2021beyond} and our method is comparable, our enhancement model is learned without low/normal-light image pairs jointly with the object detector using bounding box annotation only, which is potentially a significant advantage, \textit{e.g.} in dynamic scenes. Finally, our proposed method can be used with different detectors.

\begin{table}[!h]
\footnotesize
\begin{center}
\begin{tabular}{c c c c c c}
\toprule
 & & Proposed Enh. &  &  &  \\
Detector & init. & Module & $AP_{50}$ & $AP_{75}$ & $AP$ \\
\hline
\multirow{2}{*}{RetinaNet \cite{lin2017focal}} & \multirow{2}{*}{COCO \cite{COCO}} 
 & \xmark & 72.8\% & 47.6\% & 45.9\% \\
 & & \checkmark &\textbf{ 74.1\%} & \textbf{49.1\%} &\textbf{ 47.5\%} \\
 \hline
\multirow{2}{*}{PAA \cite{paa-eccv2020}} & \multirow{2}{*}{COCO \cite{COCO}}  
 & \xmark & 71.6\% & 47.5\% & 45.4\% \\
 & & \checkmark & \textbf{73.0\%} &\textbf{ 48.5\%} & \textbf{46.7\%} \\
 \hline
\multirow{2}{*}{Faster R-CNN \cite{Ren_2017}} & \multirow{2}{*}{ImageNet \cite{deng2009imagenet}} 
 & \xmark & 61.6\% & 34.5\% & 33.3\% \\
 & & \checkmark &\textbf{ 64.2\% }& \textbf{35.8\%} &\textbf{ 35.5\%} \\
 \hline
\multirow{2}{*}{FCOS \cite{tian2019fcos}} & \multirow{2}{*}{random} 
 & \xmark & 56.2\% & \textbf{22.1\%} & 26.4\% \\
 & & \checkmark & \textbf{58.1\%} & {21.0\%} & \textbf{26.7\%} \\
 \hline
\toprule
\end{tabular}
\vspace{-1.em}
\end{center}
\caption{Results of the proposed method, including proposed augmentation methods, on subsets of our dataset (Sony, Nikon), our dataset (Sony+Nikon) and ExDark \cite{Exdark}. In all experiments, we use RetinaNet \cite{lin2017focal} as the object detection framework. The models were trained and tested on the Sony subset. }
\label{tab:rebuttal-detectors}
\vspace{-1.1em}
\end{table}

In Tab. \ref{tab:rebuttal-detectors}, we show that our proposed method, in addition to a single-stage anchor-based detector (RetinaNet \cite{lin2017focal}), can work for other detection models as well: two-stage Faster R-CNN \cite{Ren_2017}, anchor-free FCOS \cite{tian2019fcos}, and a single-stage detector with an alternative anchor assignment scheme PAA \cite{paa-eccv2020}.

\vspace{-1.2em}\section{Conclusion}
\vspace{-0.6em}\label{sec:conclusion}
In this paper, we presented a high-quality large-scale dataset for object detection under low-light conditions showing outdoor scenes with all common challenges of low light photography: motion blur, out-of-focus blur, and noise. Further, we linked perceptual difficulty to low-light conditions and annotated instances in the Sony test set as \textit{extreme} and \textit{non-extreme}, allowing for more in-depth evaluation of methods targeting low-light conditions in the future. We expect that this dataset will be a valuable resource for the researchers in domains of object detection, low-light image enhancement, and domain adaptation, to name a few.
    
Moreover, we proposed to incorporate an image enhancement module into the object detection framework that, paired with the proposed block shuffle and patch-wise light augmentation, led to improvements over the baseline model on low-light datasets. Performance gains introduced by our method were slight but consistent -- in this paper we show that the perception under extreme low-light conditions is a difficult problem that should be addressed on its own, rather than merely a substask of object detection. 

All in all, in our paper, we highlighted that there exists a significant difficulty for object detectors under low-light conditions. In particular, we showed that there is a large performance gap under extreme and non-extreme low-light conditions that cannot be eliminated by including large amounts of extreme examples in the training or our proposed enhancement module. Paired with the observation that ConvNets do not learn to normalize features with respect to the lighting type, this suggests that machine cognition under low-light conditions is a non-trivial problem that requires special attention from researchers.

\vspace{-1.2em}
\section*{Acknowledgement}
This work was supported in part by the Ministry of Science and Technology, Taiwan, under Grant MOST 110-2634-F-002-026 and Qualcomm Technologies, Inc. We are grateful to the National Center for High-performance Computing.

\clearpage

\bibliography{llodbib}

\begin{thebibliography}{28}
\providecommand{\natexlab}[1]{#1}
\providecommand{\url}[1]{\texttt{#1}}
\expandafter\ifx\csname urlstyle\endcsname\relax
  \providecommand{\doi}[1]{doi: #1}\else
  \providecommand{\doi}{doi: \begingroup \urlstyle{rm}\Url}\fi

\bibitem[Chen et~al.(2018)Chen, Chen, Xu, and Koltun]{SID}
Chen Chen, Qifeng Chen, Jia Xu, and Vladlen Koltun.
\newblock Learning to see in the dark.
\newblock In \emph{2018 IEEE/CVF Conference on Computer Vision and Pattern
  Recognition}, pages 3291--3300, June 2018.
\newblock \doi{10.1109/CVPR.2018.00347}.

\bibitem[Chen et~al.(2019{\natexlab{a}})Chen, Chen, Do, and
  Koltun]{chen2019seeing}
Chen Chen, Qifeng Chen, Minh~N Do, and Vladlen Koltun.
\newblock Seeing motion in the dark.
\newblock In \emph{Proceedings of the IEEE/CVF International Conference on
  Computer Vision}, pages 3185--3194, 2019{\natexlab{a}}.

\bibitem[Chen et~al.(2019{\natexlab{b}})Chen, Wang, Pang, Cao, Xiong, Li, Sun,
  Feng, Liu, Xu, Zhang, Cheng, Zhu, Cheng, Zhao, Li, Lu, Zhu, Wu, Dai, Wang,
  Shi, Ouyang, Loy, and Lin]{mmdetection}
Kai Chen, Jiaqi Wang, Jiangmiao Pang, Yuhang Cao, Yu~Xiong, Xiaoxiao Li,
  Shuyang Sun, Wansen Feng, Ziwei Liu, Jiarui Xu, Zheng Zhang, Dazhi Cheng,
  Chenchen Zhu, Tianheng Cheng, Qijie Zhao, Buyu Li, Xin Lu, Rui Zhu, Yue Wu,
  Jifeng Dai, Jingdong Wang, Jianping Shi, Wanli Ouyang, Chen~Change Loy, and
  Dahua Lin.
\newblock {MMDetection}: Open mmlab detection toolbox and benchmark.
\newblock \emph{arXiv preprint arXiv:1906.07155}, 2019{\natexlab{b}}.

\bibitem[Deng et~al.(2009)Deng, Dong, Socher, Li, Li, and
  Fei-Fei]{deng2009imagenet}
Jia Deng, Wei Dong, Richard Socher, Li-Jia Li, Kai Li, and Li~Fei-Fei.
\newblock Imagenet: A large-scale hierarchical image database.
\newblock In \emph{2009 IEEE conference on computer vision and pattern
  recognition}, pages 248--255. Ieee, 2009.

\bibitem[Guo et~al.(2020)Guo, Li, Guo, Loy, Hou, Kwong, and Runmin]{Zero-DCE}
Chunle Guo, Chongyi Li, Jichang Guo, Chen~Change Loy, Junhui Hou, Sam Kwong,
  and Cong Runmin.
\newblock Zero-reference deep curve estimation for low-light image enhancement.
\newblock \emph{CVPR}, 2020.

\bibitem[Guo et~al.(2017)Guo, Li, and Ling]{LIME}
Xiaojie Guo, Yu~Li, and Haibin Ling.
\newblock Lime: Low-light image enhancement via illumination map estimation.
\newblock \emph{IEEE Transactions on Image Processing}, 26\penalty0
  (2):\penalty0 982--993, 2017.
\newblock \doi{10.1109/TIP.2016.2639450}.

\bibitem[Kim and Lee(2020)]{paa-eccv2020}
Kang Kim and Hee~Seok Lee.
\newblock Probabilistic anchor assignment with iou prediction for object
  detection.
\newblock In \emph{ECCV}, 2020.

\bibitem[Kingma and Ba(2014)]{adam}
Diederik~P Kingma and Jimmy Ba.
\newblock Adam: A method for stochastic optimization.
\newblock \emph{arXiv preprint arXiv:1412.6980}, 2014.

\bibitem[Ledig et~al.(2017)Ledig, Theis, Husz{\'a}r, Caballero, Cunningham,
  Acosta, Aitken, Tejani, Totz, Wang, et~al.]{vggloss}
Christian Ledig, Lucas Theis, Ferenc Husz{\'a}r, Jose Caballero, Andrew
  Cunningham, Alejandro Acosta, Andrew Aitken, Alykhan Tejani, Johannes Totz,
  Zehan Wang, et~al.
\newblock Photo-realistic single image super-resolution using a generative
  adversarial network.
\newblock In \emph{Proceedings of the IEEE conference on computer vision and
  pattern recognition}, pages 4681--4690, 2017.

\bibitem[Liang et~al.(2021)Liang, Wang, Quan, Chen, Liu, Ling, and Xu]{REG}
Jinxiu Liang, Jingwen Wang, Yuhui Quan, Tianyi Chen, Jiaying Liu, Haibin Ling,
  and Yong Xu.
\newblock Recurrent exposure generation for low-light face detection.
\newblock \emph{IEEE Transactions on Multimedia}, pages 1--1, 2021.
\newblock \doi{10.1109/TMM.2021.3068840}.

\bibitem[Lim et~al.(2015)Lim, Kim, Sim, and Kim]{lim2015robust}
Jaemoon Lim, Jin-Hwan Kim, Jae-Young Sim, and Chang-Su Kim.
\newblock Robust contrast enhancement of noisy low-light images:
  Denoising-enhancement-completion.
\newblock In \emph{2015 IEEE International Conference on Image Processing
  (ICIP)}, pages 4131--4135. IEEE, 2015.

\bibitem[Lin et~al.(2014)Lin, Maire, Belongie, Hays, Perona, Ramanan,
  Doll{\'a}r, and Zitnick]{COCO}
Tsung-Yi Lin, Michael Maire, Serge Belongie, James Hays, Pietro Perona, Deva
  Ramanan, Piotr Doll{\'a}r, and C~Lawrence Zitnick.
\newblock Microsoft coco: Common objects in context.
\newblock In \emph{European conference on computer vision}, pages 740--755.
  Springer, 2014.

\bibitem[Lin et~al.(2017)Lin, Goyal, Girshick, He, and
  Doll{\'a}r]{lin2017focal}
Tsung-Yi Lin, Priya Goyal, Ross Girshick, Kaiming He, and Piotr Doll{\'a}r.
\newblock Focal loss for dense object detection.
\newblock In \emph{Proceedings of the IEEE international conference on computer
  vision}, pages 2980--2988, 2017.

\bibitem[Loh and Chan(2019)]{Exdark}
Yuen~Peng Loh and Chee~Seng Chan.
\newblock Getting to know low-light images with the exclusively dark dataset.
\newblock \emph{Computer Vision and Image Understanding}, 178:\penalty0 30--42,
  2019.
\newblock \doi{https://doi.org/10.1016/j.cviu.2018.10.010}.

\bibitem[Lore et~al.(2017)Lore, Akintayo, and Sarkar]{LORE2017650}
Kin~Gwn Lore, Adedotun Akintayo, and Soumik Sarkar.
\newblock Llnet: A deep autoencoder approach to natural low-light image
  enhancement.
\newblock \emph{Pattern Recognition}, 61:\penalty0 650--662, 2017.
\newblock ISSN 0031-3203.
\newblock \doi{https://doi.org/10.1016/j.patcog.2016.06.008}.
\newblock URL
  \url{https://www.sciencedirect.com/science/article/pii/S003132031630125X}.

\bibitem[Lv et~al.(2018)Lv, Lu, Wu, and Lim]{lv2018mbllen}
Feifan Lv, Feng Lu, Jianhua Wu, and Chongsoon Lim.
\newblock Mbllen: Low-light image/video enhancement using cnns.
\newblock In \emph{BMVC}, page 220, 2018.

\bibitem[Misra(2019)]{mish}
Diganta Misra.
\newblock Mish: A self regularized non-monotonic activation function.
\newblock \emph{arXiv preprint arXiv:1908.08681}, 2019.

\bibitem[Ratnasingam(2019)]{ratnasingam2019deep}
Sivalogeswaran Ratnasingam.
\newblock Deep camera: A fully convolutional neural network for image signal
  processing.
\newblock In \emph{Proceedings of the IEEE/CVF International Conference on
  Computer Vision Workshops}, pages 0--0, 2019.

\bibitem[Ren et~al.(2017)Ren, He, Girshick, and Sun]{Ren_2017}
Shaoqing Ren, Kaiming He, Ross Girshick, and Jian Sun.
\newblock Faster r-cnn: Towards real-time object detection with region proposal
  networks.
\newblock \emph{IEEE Transactions on Pattern Analysis and Machine
  Intelligence}, Jun 2017.

\bibitem[Ronneberger et~al.(2015)Ronneberger, Fischer, and Brox]{unet}
Olaf Ronneberger, Philipp Fischer, and Thomas Brox.
\newblock U-net: Convolutional networks for biomedical image segmentation.
\newblock In \emph{International Conference on Medical image computing and
  computer-assisted intervention}, pages 234--241. Springer, 2015.

\bibitem[Schwartz et~al.(2018)Schwartz, Giryes, and
  Bronstein]{schwartz2018deepisp}
Eli Schwartz, Raja Giryes, and Alex~M Bronstein.
\newblock Deepisp: Toward learning an end-to-end image processing pipeline.
\newblock \emph{IEEE Transactions on Image Processing}, 28\penalty0
  (2):\penalty0 912--923, 2018.

\bibitem[Tian et~al.(2019)Tian, Shen, Chen, and He]{tian2019fcos}
Zhi Tian, Chunhua Shen, Hao Chen, and Tong He.
\newblock Fcos: Fully convolutional one-stage object detection.
\newblock \emph{arXiv preprint arXiv:1904.01355}, 2019.

\bibitem[Wang et~al.(2019)Wang, Zhang, Fu, Shen, Zheng, and
  Jia]{wang2019underexposed}
Ruixing Wang, Qing Zhang, Chi-Wing Fu, Xiaoyong Shen, Wei-Shi Zheng, and Jiaya
  Jia.
\newblock Underexposed photo enhancement using deep illumination estimation.
\newblock In \emph{Proceedings of the IEEE/CVF Conference on Computer Vision
  and Pattern Recognition}, pages 6849--6857, 2019.

\bibitem[Wang et~al.(2004)Wang, Bovik, Sheikh, and Simoncelli]{ssim}
Zhou Wang, A.C. Bovik, H.R. Sheikh, and E.P. Simoncelli.
\newblock Image quality assessment: from error visibility to structural
  similarity.
\newblock \emph{IEEE Transactions on Image Processing}, 13\penalty0
  (4):\penalty0 600--612, 2004.
\newblock \doi{10.1109/TIP.2003.819861}.

\bibitem[Wei et~al.(2018)Wei, Wang, Yang, and Liu]{wei2018deep}
Chen Wei, Wenjing Wang, Wenhan Yang, and Jiaying Liu.
\newblock Deep retinex decomposition for low-light enhancement.
\newblock \emph{arXiv preprint arXiv:1808.04560}, 2018.

\bibitem[Yang et~al.(2020)Yang, Yuan, Ren, Liu, Scheirer, Wang, Zhang, Zhong,
  Xie, Pu, Zheng, Qu, Xie, Chen, Li, Hong, Jiang, Yang, Liu, Qu, Wan, Zheng,
  Zhong, Su, He, Guo, Zhao, Zhu, Liang, Wang, Chen, Quan, Xu, Liu, Liu, Sun,
  Lin, Li, Lu, Gu, Zhou, Cao, Zhang, Chi, Zhuang, Lei, Li, Wang, Liu, Yi, Zuo,
  Chi, Wang, Wang, Liu, Gao, Chen, Guo, Li, Zhong, Huang, Guo, Yang, Liao,
  Yang, Zhou, Feng, and Qin]{poor_visibility_benchmark}
Wenhan Yang, Ye~Yuan, Wenqi Ren, Jiaying Liu, Walter~J. Scheirer, Zhangyang
  Wang, Taiheng Zhang, Qiaoyong Zhong, Di~Xie, Shiliang Pu, Yuqiang Zheng,
  Yanyun Qu, Yuhong Xie, Liang Chen, Zhonghao Li, Chen Hong, Hao Jiang, Siyuan
  Yang, Yan Liu, Xiaochao Qu, Pengfei Wan, Shuai Zheng, Minhui Zhong, Taiyi Su,
  Lingzhi He, Yandong Guo, Yao Zhao, Zhenfeng Zhu, Jinxiu Liang, Jingwen Wang,
  Tianyi Chen, Yuhui Quan, Yong Xu, Bo~Liu, Xin Liu, Qi~Sun, Tingyu Lin,
  Xiaochuan Li, Feng Lu, Lin Gu, Shengdi Zhou, Cong Cao, Shifeng Zhang, Cheng
  Chi, Chubing Zhuang, Zhen Lei, Stan~Z. Li, Shizheng Wang, Ruizhe Liu, Dong
  Yi, Zheming Zuo, Jianning Chi, Huan Wang, Kai Wang, Yixiu Liu, Xingyu Gao,
  Zhenyu Chen, Chang Guo, Yongzhou Li, Huicai Zhong, Jing Huang, Heng Guo,
  Jianfei Yang, Wenjuan Liao, Jiangang Yang, Liguo Zhou, Mingyue Feng, and
  Likun Qin.
\newblock Advancing image understanding in poor visibility environments: A
  collective benchmark study.
\newblock \emph{IEEE Transactions on Image Processing}, 29:\penalty0
  5737--5752, 2020.
\newblock \doi{10.1109/TIP.2020.2981922}.

\bibitem[Zhang et~al.(2019)Zhang, Zhang, and Guo]{zhang2019kindling}
Yonghua Zhang, Jiawan Zhang, and Xiaojie Guo.
\newblock Kindling the darkness: A practical low-light image enhancer.
\newblock In \emph{Proceedings of the 27th ACM International Conference on
  Multimedia}, pages 1632--1640, 2019.

\bibitem[Zhang et~al.(2021)Zhang, Guo, Ma, Liu, and Zhang]{zhang2021beyond}
Yonghua Zhang, Xiaojie Guo, Jiayi Ma, Wei Liu, and Jiawan Zhang.
\newblock Beyond brightening low-light images.
\newblock \emph{International Journal of Computer Vision}, 129\penalty0
  (4):\penalty0 1013--1037, 2021.

\end{thebibliography}
\end{document}